\documentclass[letterpaper, 10 pt, conference]{ieeeconf}  

\IEEEoverridecommandlockouts                              

\pdfoutput=1
\usepackage{graphics} 
\usepackage{epsfig} 
\usepackage{amsmath} 
\usepackage{amssymb}  
\usepackage{epstopdf}
\usepackage{graphicx}
\usepackage{subfigure}
\usepackage{color}
\usepackage{cite}
\usepackage{url}
\usepackage{bigstrut}

\title{\LARGE \bf
Dynamic Object Tracking for Quadruped Manipulator with \\
Spherical Image-Based Approach
}

\author{Tianlin Zhang, 
 Sikai Guo,
Xiaogang Xiong,
Wanlei Li,
Zezheng Qi,
and
Yunjiang Lou
\thanks{*This work was supported partially by the National Key Research and Development Program of China under Grant No. 2020YFB1313900, Shenzhen Fundamental Research
Program Grants KQTD20190929172545139,
JSGG20210420091602008 and partially supported by GuangDong Research
Foundation 2022A1515011521, (Corresponding authors: Xiaogang Xiong and Yunjiang Lou)}
\thanks{T. Zhang, S. Guo, X. Xiong, W. Li, Z. Qi and Y. Lou are with the School of Mechanical Engineering and Automation,
 Harbin Institute of Technology, Shenzhen, 518000, P.R. China (e-mail: skywoodszcn@gmail.com; 200320818@stu.hit.edu.cn; xiongxg@hit.edu.cn; 19b953034@stu.hit.edu.cn; qizezheng@stu.hit.edu.cn; louyj@hit.edu.cn).}%
}

\begin{document}

\maketitle
\thispagestyle{empty}
\pagestyle{empty}

\begin{abstract}
Exactly estimating and tracking the motion of surrounding dynamic objects is one of important tasks for the autonomy of a quadruped manipulator.
However, with only an onboard RGB camera, it is still a challenging work for a quadruped manipulator to track the motion of a dynamic object moving with unknown and changing velocities.
To address this problem, this manuscript proposes a novel image-based visual servoing (IBVS) approach consisting of three
elements: a spherical projection model, a robust super-twisting observer, and a model predictive controller (MPC).
The spherical projection model decouples the visual error of the dynamic target into linear and angular ones.
Then, with the presence of the visual error, the robustness of the observer is exploited to estimate the unknown and changing velocities of the dynamic target without depth estimation.
Finally, the estimated velocity is fed into the model predictive controller (MPC) to generate joint torques for the quadruped manipulator to track the motion of the dynamical target.
The proposed approach is validated through hardware experiments and the experimental results illustrate the approach's effectiveness in improving the autonomy of the quadruped manipulator.
\end{abstract}

\section{INTRODUCTION}
\begin{figure}[tp]
  \centering
  \includegraphics[scale=0.38]{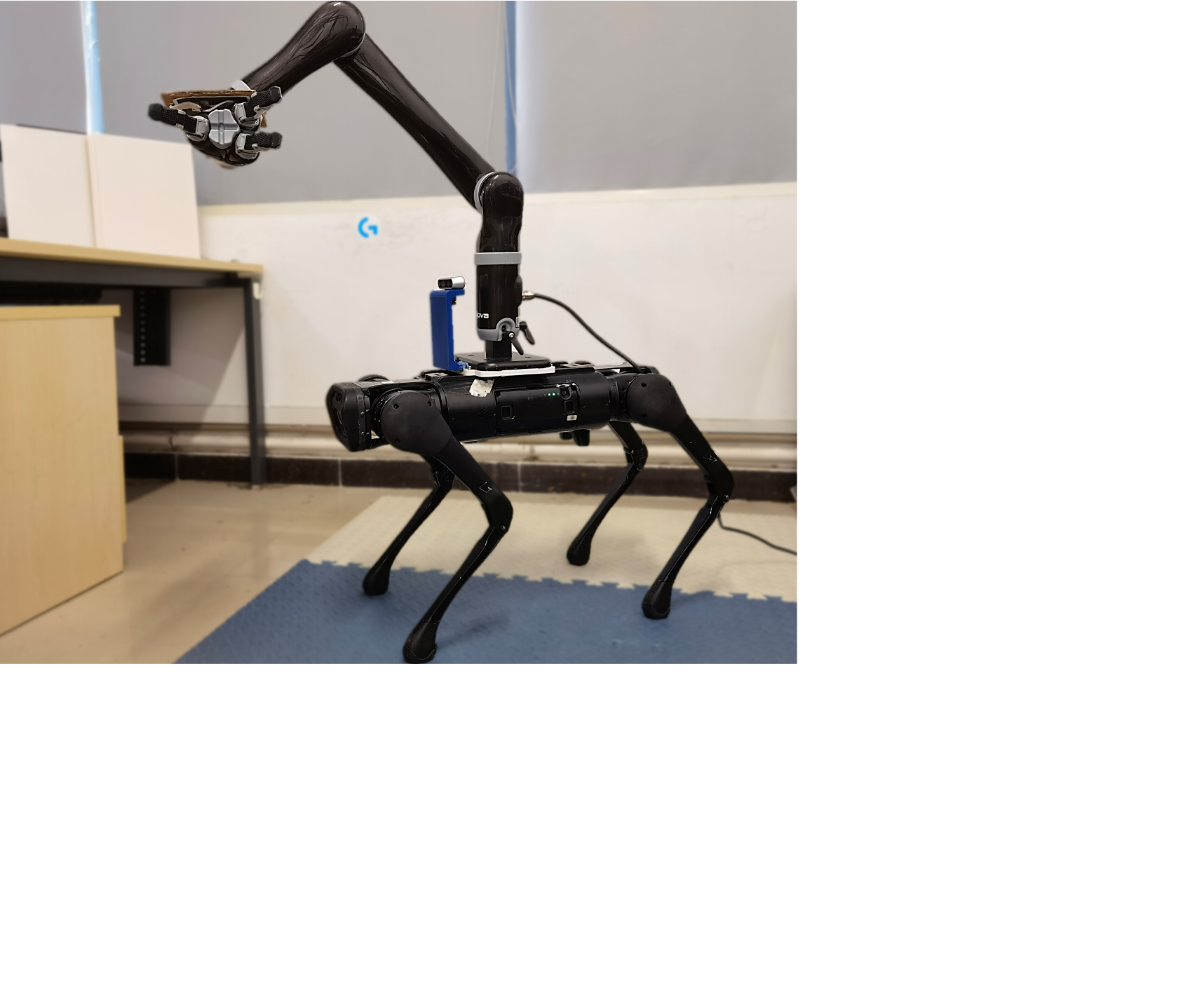}
  \caption{An illustration of a quadruped manipulator consisted of an Aliengo platform of quadruped robot and one 6-DOF Kinova manipulator for grasping surrounding objects.}
  \label{fig:qm_grasp}
\end{figure}

The quadruped manipulator is a multi-functional platform that is comprised of a mobile quadruped and a manipulator, as shown in Fig.~\ref{fig:qm_grasp}. It has not only the agility of the quadruped but also the interactivity of the manipulator, and thus has great potential applications, such as disaster rescue\cite{biswal2021development} and anti-terrorist operations\cite{Li2011}. To exploit these advances, making the quadruped manipulator to track the motion of surrounding dynamic objects is an important task for the autonomy of the quadruped manipulator.


In the previous studies \cite{Rehman2016a,Bellicoso2019a,Ferrolho2020,Sleiman2021,Chiu2022a}, external humans or marker boards (e.g., ArUco) are needed to provide the motion expectations to the quadruped manipulator.
They assumed that the motion state of the target to be track was known and static, which limits the scope of the applications.
More notable contributions to the autonomy of the quadruped manipulator could be found in the works\cite{zimmermann2021go ,Mittal2021}, where the target state was estimated with onboard RGB-D cameras, and motion expectations of the robot were automatically generated through trajectory optimization.
However, they only solved the problem of grasping static objects and had to use the depth of the target, which is often computationally expensive to obtain or inaccurate due to the environment changes in practice\cite{Ming2021}.
Dynamic target tracking, i.e., tracking an object with unknown and changing speeds, is still a difficult task for the quadruped manipulator.

Visual servoing is one of the promising methods that can enhance the autonomy of the quadruped manipulator, which has been developed for a long time and has been successfully used in various robots, such as drones \cite{Rafique2020}, wheeled robots \cite{Ye2015}, and quadrupeds \cite{kolter2009stereo}.
However, directly applying the conventional visual servoing to the quadruped manipulator has at least two challenges to solve.
The first challenge is to deal with the coupling motion between the locomotion and manipulation,
which requires the visual servoing to take the full system states into account.
The work in \cite{Chen2019} considered the locomotion and manipulation as two separate subsystems.
However, without considering the motion coupling,
it is difficult to exploit the coordination of locomotion and manipulation for the quadruped manipulator, limiting the robot's motion agility.
The second challenge is caused by the under-actuation of the quadruped platform.
When the robot operates under a dynamic gait (e.g., trot),
the roll and pitch angles of the quadruped platform is uncontrollable due to the under-actuation,
which seriously affects the operation accuracy of the manipulator's end-effector.

The above two challenges of the visual servoing were also studied for aerial robots in the literature.
For example, the work in \cite{Hamel2002,Odile2007a,Guenard2008} employed the image-based visual seroving (IBVS) to track a static target but do not consider the coupling of the manipulator and the quadrotor.
In contrast,
Zhong et al. \cite{Zhong2020} considered the coupling between the manipulator and the aerial drone but only tracked a static target with the aerial manipulator.
To track a dynamic object, in \cite{Mahony2008,Herisse2012,Serra2016}, the optical flow technology was adopted to estimate the velocity of the dynamic target.
The problem is that, when the target depth is unknown, the optical flow can not estimate the target's velocity, and it is unreliable under the assumption of constant brightness.

To track a dynamic target for the quadruped manipulator without depth estimation, this manuscript proposes a novel image-based visual servoing (IBVS) approach that consists of three
elements: a spherical projection model, a robust super-twisting observer (STO), and a model predictive controller (MPC).
First, by simultaneously projecting the images of the dynamic target and the manipulator onto the spherical image plane, we have the equation of visual error with the passive-like property for the quadruped manipulator.
The passivity-like property of the visual error allows us to separately control the position and attitude of the under-actuated quadruped manipulator, which can avoid the influence of the quadruped platform to the attitude on error convergence \cite{Zhong2020}.
 Then, with the spherical projection model, the robustness of the multi-variable STO \cite{Yang2022} is exploited to estimate the target's velocity without depth estimation in spite of the presence of visual error.
Finally, to track the desired motion, by adopting the work \cite{Rehman2016a,DiCarlo2018}, a model predictive controller (MPC) for the quadruped manipulator based on a model of single rigid body is used to generate joint torques.

This work offers the following contributions:
\begin{itemize}
    \item We propose a new spherical image-based approach to enable the quadruped manipulator to
    exactly estimate the unknown velocity of a dynamic object, without the need for marker boards or depth estimation.
    \item The passive-like property of visual error is exploited for the quadruped manipulator to eliminate the influences of the under-actuated quadruped platform on the visual observation, which enable the manipulator to track the dynamic object with simultaneous locomotion.
   \item The proposed solution of dynamic object tracking for the quadruped manipulator has been verified through simulations and hardware experiments.
\end{itemize}
To the best of the authors' knowledge, this is the first time to solve the problem of tracking the dynamic object for the quadruped manipulator.

%
\begin{figure}[tp]
  \centering
  \includegraphics[scale=0.43]{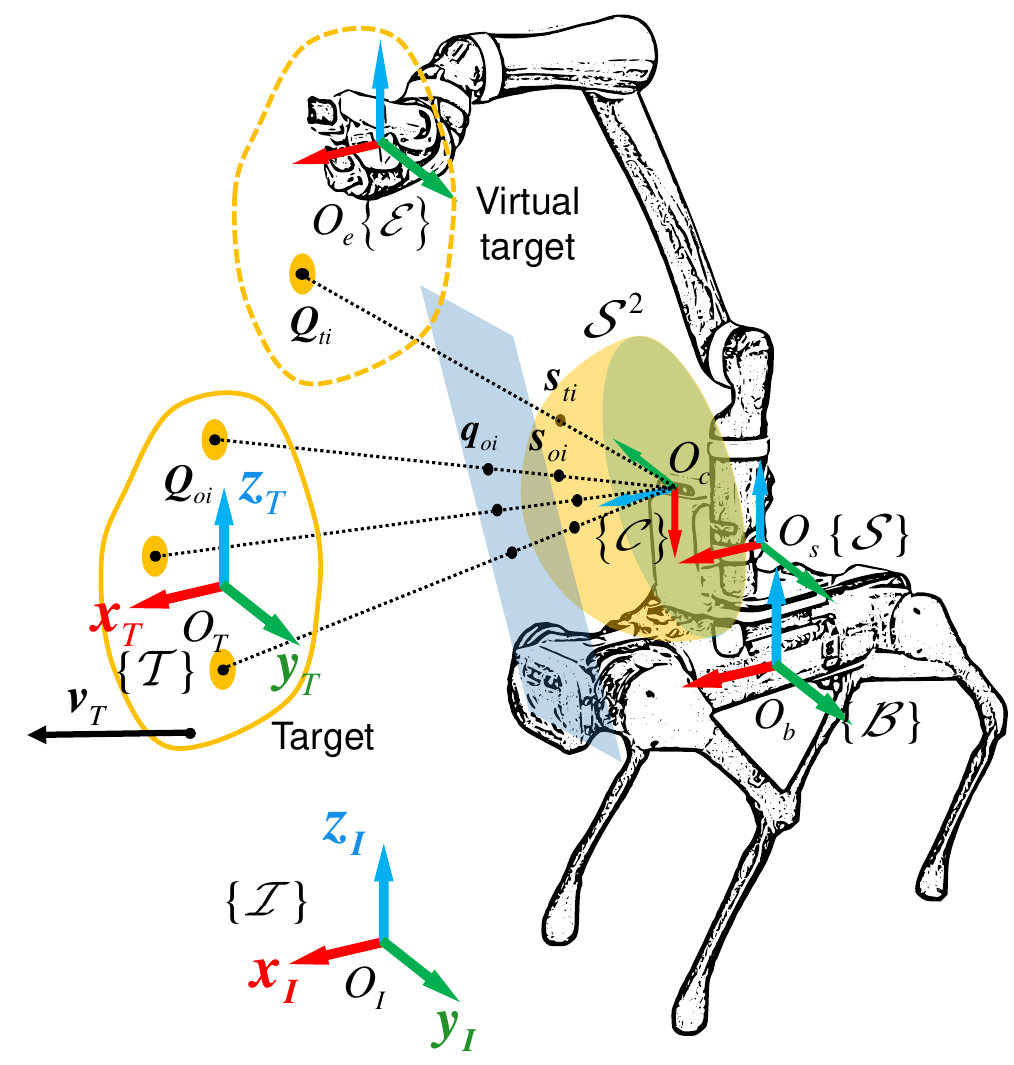}
  \caption{Quadruped manipulator reference frames.
  The blue plane represents the normalized focal length imaging plane of the camera, and the yellow sphere represents the projection sphere.}
  \label{fig:qm_frame}
\end{figure}
\section{PROBLEM FORMULATION}
Let us consider the visual tracking system that includes a  quadruped manipulator with a monocular camera and a dynamic object with unknown changing velocity, as shown in Fig. \ref{fig:qm_frame}, where
$\left\{ {{\cal I}:{O_I} - {{\boldsymbol{x}}_I}{{\boldsymbol{y}}_I}{{\boldsymbol{z}}_I}} \right\}$  represents the right-hand inertial coordinate frame, $\boldsymbol{z}_I$ denote the inverse direction of gravity, and
$\left\{ {{\cal B}:{O_b} - {{\boldsymbol{x}}_b}{{\boldsymbol{y}}_b}{{\boldsymbol{z}}_b}} \right\}$ represents the body frame of the quadruped platform with ${O_b}$ located at the centroid of the quadruped platform.
Let $\left\{ {{\cal S}:{O_s} - {{\boldsymbol{x}}_s}{{\boldsymbol{y}}_s}{{\boldsymbol{z}}_s}} \right\}$ denote the base frame of the manipulator, which is attached to the top of the quadruped platform,
and $\left\{ {{\cal E}:{O_e} - {{\boldsymbol{x}}_e}{{\boldsymbol{y}}_e}{{\boldsymbol{z}}_e}} \right\}$ represent the end-effector frame of the manipulator.
Then, $\left\{ {{\cal C}:{O_c} - {{\boldsymbol{x}}_c}{{\boldsymbol{y}}_c}{{\boldsymbol{z}}_c}} \right\}$ denotes the camera frame, where
$\boldsymbol{z}_c$  is along the optical axis, and ${{\cal S}^2}$  is the sphere normalized imaging plane.
The target image features and the manipulator's end-effector will be projected on ${{\cal S}^2}$ to describe the motion.
Let ${{\boldsymbol{p}}_{B}} \in {\mathbb{R}^3}$ represent the position of the body frame ${\cal B}$ w.r.t. the inertial coordinate frame ${\cal I}$.
The orientation of the body frame ${\cal B}$ w.r.t. the inertial coordinate frame ${\cal I}$ is parameterized using ZYX-Euler angles ${{\boldsymbol{\Phi }}_{B}}$.
Let ${{\boldsymbol{q}}_j} \in {\mathbb{R}^{{18}}}$ be the limb joint positions.
Therefore, the generalized coordinate vector ${\boldsymbol{q}}$ and the generalized velocity vector ${\boldsymbol{v}}$ are written as
\begin{equation}
\label{eq:model}
{\boldsymbol{q}} = \left[ {\begin{array}{*{20}{c}}
  {{{\boldsymbol{p}}_{B}}} \\
  {{{\boldsymbol{\Phi }}_{B}}} \\
  {{{\boldsymbol{q}}_j}}
\end{array}} \right] \in SE(3) \times {\mathbb{R}^{{18}}}{\text{, }}
{\boldsymbol{v}} = \left[ {\begin{array}{*{20}{c}}
  {{{\boldsymbol{v}}_B}} \\
  {{{\boldsymbol{\omega }}_{B}}} \\
  {{{{\boldsymbol{\dot q}}}_j}}
\end{array}} \right] \in {\mathbb{R}^{{24}}}
\end{equation}

Let $\left\{ {{\cal T}:{O_T} - {{\boldsymbol{x}}_T}{{\boldsymbol{y}}_T}{{\boldsymbol{z}}_T}} \right\}$ denote the target frame, where the target frame's attitude is aligned with the inertial coordinate frame. Let ${({{\boldsymbol{v}}_T},{{\boldsymbol{\omega }}_T})^ \top }$ be the unknown linear and angular velocities of the target frame, respectively.
To simplify the controller, the following assumptions are made related to the unknown target:
The target is translational such that ${{\boldsymbol{v}}_T} \ne 0$ and ${{\boldsymbol{v}}_T}$ is unknown. Its orientation is approximately constant such that ${{\boldsymbol{\omega }}_T} = 0$,
 and the linear acceleration ${{{\boldsymbol{\dot v}}}_T}$ is bounded.

Specifically, with the quadruped manipulator equipped with only an onboard RGB camera, the goal is first to exactly estimate the unknown and changing velocities of the target without depth estimation and any external sensors, and then to design a controller for the quadruped manipulator so that the origin of the manipulator's end-effector
$O_e$  can coincide with the target origin $O_T$.

\section{MOTION TRACKING of QUADRUPED MANIPULATOR}
\subsection{Spherical Projection Model}
Let ${{\boldsymbol{Q}}_{oi}}, i = 1, \cdots ,m$ be the feature points located on the target, expressed in the camera frame $\cal C$, as shown in Fig. \ref{fig:qm_frame}.
To build an equation of visual error with the passive-like property, these feature points are projected onto the sphere ${{\cal S}^2}$. From \cite{Hamel2002}\cite{Zhong2020}, the spherical projection of the visible image points can be numerically computed as:
\begin{equation}
\label{eq:SphericalProjection}
  \boldsymbol{s}_{o i}=\frac{\boldsymbol{q}_{o i}}{\left|\boldsymbol{q}_{o i}\right|}=\frac{\boldsymbol{Q}_{o i}}{\left|\boldsymbol{Q}_{o i}\right|}
\end{equation}
where $\boldsymbol{s}_{o i} \in \mathbb{R}^{3}$ is the vector describing target feature points being projected on the sphere ${\mathcal{S}^2}$, $| \cdot |$ is the Euclidean distance operator, and
$\boldsymbol{q}_{o i}$ is the point expressed in the normalized focal length imaging plane.
The relationship between $\boldsymbol{s}_{o i}$, $\boldsymbol{q}_{o i}$ and $\boldsymbol{Q}_{o i}$ is shown in Fig. \ref{fig:qm_frame}.
Notes that $\boldsymbol{Q}_{o i}$  has the same linear velocity 
 as the target, and thus, based on the method in \cite{Herisse2012,Serra2016}, the kinematic equation of
$\boldsymbol{s}_{o i}$  can be written as
\begin{equation}
\label{eq:KinematicSphere}
{{{\boldsymbol{\dot s}}}_{oi}} =  - {\left[ {{\boldsymbol{\Omega }}_c^c} \right]_ \times }{{\boldsymbol{s}}_{oi}} - \frac{{{{\boldsymbol{\pi }}_{{s_{oi}}}}}}{{r\left( {{{\boldsymbol{Q}}_{oi}}} \right)}}({\boldsymbol{v}}_c^c - {\boldsymbol{v}}_T^c)
\end{equation}
where ${{\boldsymbol{\pi }}_{{s_{oi}}}} = \left( {{\mathbb{I}_{3 \times 3}} - {{\boldsymbol{s}}_{oi}}{\boldsymbol{s}}_{oi}^ \top } \right) \in {\mathbb{R}^{3 \times 3}}$,
${\boldsymbol{v}}_c^c$ and ${\mathbf{\Omega }}_c^c$ are the linear and angular velocities of the camera w.r.t. the camera frame $\cal C$, respectively, ${[ \cdot ]_ \times }$ is the skew-matrix operator,
$r\left( {{{\boldsymbol{Q}}_{oi}}} \right) = |{{\boldsymbol{Q}}_{oi}}|$  is target's depth,
$\boldsymbol{v}_T^c$ is the target's linear velocity expressed in the camera frame $\cal C$.
Notes that $r\left( {{{\boldsymbol{Q}}_{oi}}} \right)$  and $\boldsymbol{v}_T^c$ both are unavailable.
As shown in \eqref{eq:KinematicSphere}, the rotational and translational motion of the camera is decoupled through the technique of spherical projection. Therefore, a controller that only controls the translational motion can be designed to avoid the under-actuation of the quadruped platform.

For $m$ target points, the normalized spherical centroid of $\boldsymbol{s}_{o i}$ and its derivative w.r.t time are as follows \cite{Guenard2008}:
\begin{subequations}
\label{eq:ho_ho_dot}
\begin{align}
\label{eq:ho}
&{{\boldsymbol{h}}_o} = \frac{1}{m}\sum\limits_{i = 1}^m {{{\boldsymbol{s}}_{oi}}}, \,\boldsymbol{L}_o =\dfrac{1}{m} \sum\limits_{i = 1}^m \frac{{{\boldsymbol{\pi }}_{{s_{oi}}}}}{| {{{\boldsymbol{Q}}_{oi}}}|} \\
\label{eq:ho_dot}
&{{{\boldsymbol{\dot h}}}_o} =  - {\left[ {{\boldsymbol{\Omega }}_c^c} \right]_ \times }{{\boldsymbol{h}}_o} - {\boldsymbol{L}_o}({\boldsymbol{v}}_c^c - {\boldsymbol{v}}_T^c)
\end{align}
\end{subequations}
where $\boldsymbol{L}_o \in {\mathbb{R}^{3 \times 3}}$ is a positive definite matrix when there are at least two feature points,
and ${{\boldsymbol{h}}_o}$ can represent the target center.

In order to describe the motion of the manipulator on ${{\cal S}^2}$, some virtual points ${{\boldsymbol{Q}}_{ti}}$ are captured around the center of the manipulator's end-effector frame $\cal E$, expressed in the camera frame $\cal C$, as shown in Fig. \ref{fig:qm_frame}.
Therefore,  $\boldsymbol{Q}_{ti}=\boldsymbol{O}_e+\Delta {{\boldsymbol{\rho }}_i}$,
where $\boldsymbol{O}_e$ is the origin position of the manipulator's end-effector frame ${\cal E}$,
$\Delta \boldsymbol{\rho }_i$ is the distance between ${{\boldsymbol{Q}}_{ti}}$ and $\boldsymbol{O}_e$, which can be specified manually.

\begin{figure}[tp]
  \centering
  \includegraphics[scale=0.38]{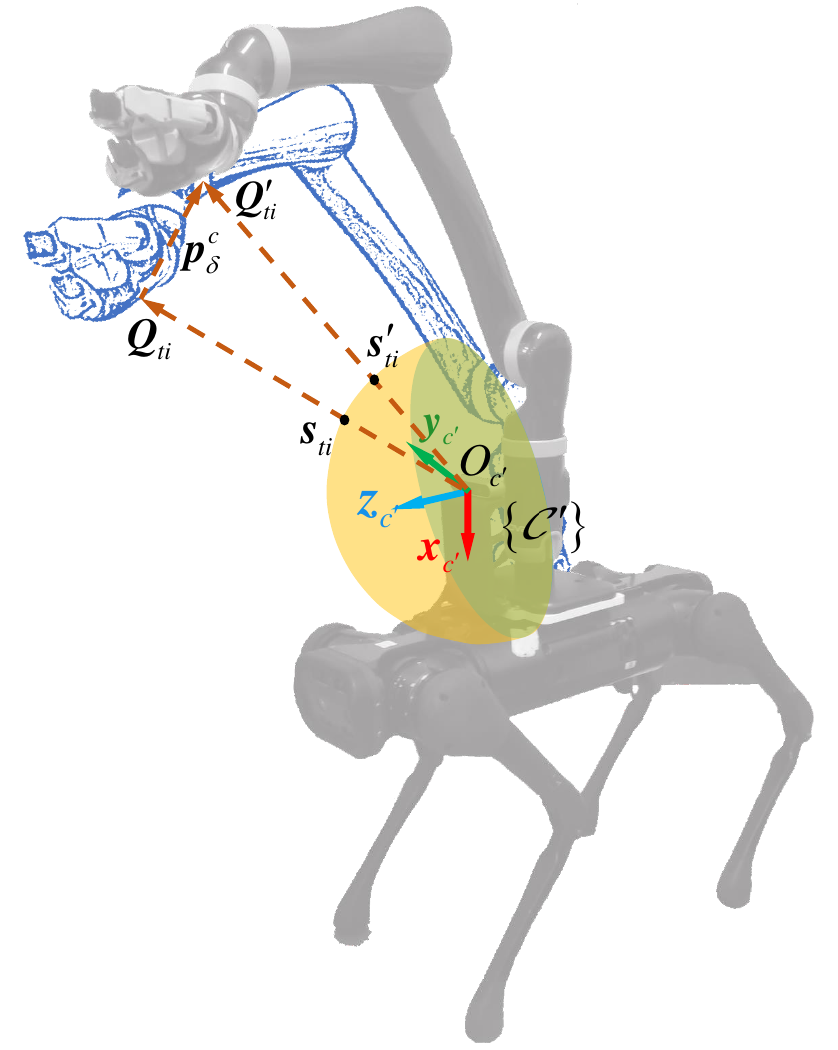}
  \caption{The representative snapshot of the movement of the real and the virtual manipulators. The blue represents the real manipulator, while the gray one represents the virtual manipulator. The feature points ${{\boldsymbol{Q}}_{ti}}$ and ${{\boldsymbol{Q}}^\prime_{ti}}$ of both manipulators are projected to the sphere coordinate that represented by the yellow color.}
  \label{fig:qm_virtual}
\end{figure}

\begin{figure*}[htp]
  \centering
  \includegraphics[scale=0.49]{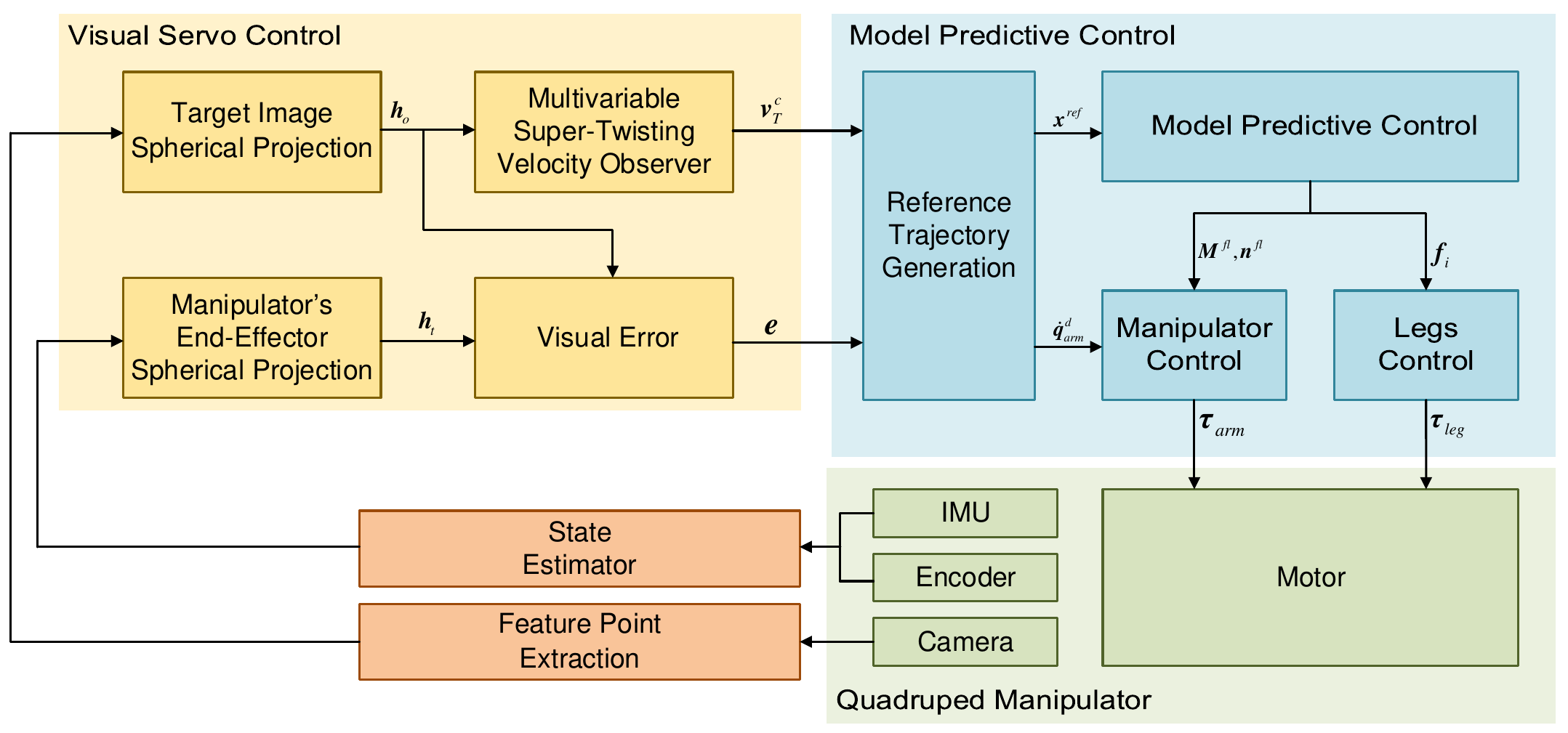}
  \caption{The block diagram of the proposed approach for the quadruped manipulator control.
  Firstly, the visual servoing control uses the robot's state and extracted feature points to construct the visual error $\boldsymbol{e}$.
  Meanwhile, the target's unknown velocity $\boldsymbol{v}_T^c$ is estimated by the multi-variable super-twisting observer (STO).
  Afterward, the model predictive control generates torque for each joint based on visual error $\boldsymbol{e}$ and estimated target velocity $\boldsymbol{v}_T^c$ from the visual servoing control.
  }
  \label{fig:qm_control}
\end{figure*}

To build an equation of visual error with a passive-like property, a virtual plane of the camera is created\cite{Rafique2020}, as shown in Fig. \ref{fig:qm_virtual}.
Let $\left\{ {{\cal C}':{O_{c'}} - {{\boldsymbol{x}}_{c'}}{{\boldsymbol{y}}_{c'}}{{\boldsymbol{z}}_{c'}}} \right\}$ denote the
virtual plane frame of the camera, which means that roll and pitch angles of the quadruped platform are assumed to be zero during the motion of the system.
Let ${{{\boldsymbol{Q'}}}_{ti}}$  denote the redefined virtual target point ${{{\boldsymbol{Q}}}_{ti}}$ in the camera's virtual plane frame ${\cal C}'$.

Therefore, ${{{\boldsymbol{Q'}}}_{ti}}$ can be calculated by
\begin{equation}
\label{eq:visual_manipulator}
\begin{aligned}
{{{\boldsymbol{Q'}}}_{ti}} &= {{\boldsymbol{Q}}_{ti}} + {\boldsymbol{p}}_\delta ^c\\
&= {{\boldsymbol{Q}}_{ti}} + {}_{c'}^c{\boldsymbol{R}}\left( {{\boldsymbol{p}}_{{{{\boldsymbol{Q'}}}_{ti}}}^{c'} - {\boldsymbol{p}}_{{{\boldsymbol{Q}}_{ti}}}^{c'}} \right)\\
&= {{\boldsymbol{Q}}_{ti}} + {}_{c'}^c{\boldsymbol{R}}\left( {\left( {{\boldsymbol{p}}_{{O_c}}^{c'} + {{\boldsymbol{Q}}_{ti}}} \right) - \left( {{\boldsymbol{p}}_{{O_c}}^{c'} + {}_c^{c'}{\boldsymbol{R}}{{\boldsymbol{Q}}_{ti}}} \right)} \right) \\
&= {}_c^{c'}{{\boldsymbol{R}}^ \top }{{\boldsymbol{Q}}_{ti}}
 \end{aligned}
\end{equation}
where $\boldsymbol{p}_\delta ^c$  is the offset vector caused by the quadruped rotation, expressed in the camera frame ${\cal C}$,
${}_c^{c'}\boldsymbol{R}$ represents the rotation matrix of the camera frame ${\cal C}$ w.r.t the camera's virtual plane frame ${\cal C}'$,
${\boldsymbol{p}}_{{{{\boldsymbol{Q'}}}_{ti}}}^{c'}$, ${\boldsymbol{p}}_{{{\boldsymbol{Q}}_{ti}}}^{c'}$, and ${\boldsymbol{p}}_{{O_c}}^{c'}$ are the offset vector of ${{\boldsymbol{Q}}_{ti}}$, ${{{\boldsymbol{Q'}}}_{ti}}$ and the origin of the camera frame ${\cal C}$, expressed in the camera's virtual plane frame ${{\cal C}'}$, respectively.
The relationship can be seen in Fig. \ref{fig:qm_virtual}.

Taking the derivative of ${{{\boldsymbol{Q'}}}_{ti}}$ w.r.t time, we can get
\begin{equation}
\label{eq:Qti_kinematics}
{{{\boldsymbol{\dot Q'}}}_{ti}} = {}_c^{c'}{{\boldsymbol{R}}^ \top }{{{\boldsymbol{\dot Q}}}_{ti}} - {\left[ {{\boldsymbol{\Omega }}_c^c} \right]_ \times }{{{\boldsymbol{Q'}}}_{ti}}.
\end{equation}
The point ${{{\boldsymbol{Q'}}}_{ti}}$ projected onto sphere plane ${{\cal S}^2}$ is expressed as ${{{\boldsymbol{s'}}}_{ti}} = \frac{{{{{\boldsymbol{Q'}}}_{ti}}}}{{\left| {{{{\boldsymbol{Q'}}}_{ti}}} \right|}}$.
Therefore,  the derivative of ${{{\boldsymbol{s'}}}_{ti}}$  w.r.t time is
%
\begin{equation}
\label{eq:sti_kinematics2}
{{{\boldsymbol{\dot s'}}}_{ti}} = \frac{{{{\boldsymbol{\pi }}_{{{{{s'}}}_{ti}}}}}}{{r\left( {{{{\boldsymbol{Q'}}}_{ti}}} \right)}}{}_c^{c'}{{\boldsymbol{R}}^ \top }{{{\boldsymbol{\dot Q}}}_{ti}} - {\left[ {{\boldsymbol{\Omega }}_c^c} \right]_ \times }{{{\boldsymbol{s'}}}_{ti}}
\end{equation}
where ${{\boldsymbol{\pi }}_{{{{{s'}}}_{ti}}}} = \left( {{\mathbb{I}_{3 \times 3}} - {{{\boldsymbol{s'}}}_{ti}}{{{\boldsymbol{s'}}}_{ti}}^ \top } \right)$.
From manipulator's kinematics, ${{{\boldsymbol{\dot Q}}}_{ti}}$ can be calculated as
\begin{equation}
\label{eq:arm_joint}
\begin{aligned}
{{{\boldsymbol{\dot Q}}}_{ti}} &= {}_s^c{\boldsymbol{R}}\left( {\begin{array}{*{20}{l}}
  {{\mathbb{I}_{3 \times 3}}}&-{{{\left[ {_e^s{\boldsymbol{R}}\Delta {{\boldsymbol{\rho }}_i}} \right]}_ \times }}
\end{array}} \right)\left( {\begin{array}{*{20}{c}}
  {_e^s{\boldsymbol{\dot t}}} \\
  {_e^s{\boldsymbol{w}}}
\end{array}} \right)\\
&={{\boldsymbol{J}}_{ti}}{{\boldsymbol{J}}_e}{\boldsymbol{Sv}}
\end{aligned}
\end{equation}
where $_e^s{\boldsymbol{\dot t}}$ and $_e^s{\boldsymbol{w}}$ are linear and angular velocity of the manipulator's end-effector frame $\cal{E}$ w.r.t the manipulator's base frame $\cal{S}$, respectively,
$_e^s{\boldsymbol{R}}$ represents the rotation matrix of the manipulator's end-effector frame $\cal{E}$  w.r.t the manipulator's base frame $\cal{S}$,
${}_s^c{\boldsymbol{R}}$ represents the rotation matrix of the manipulator's base frame $\cal{S}$  w.r.t the camera frame $\cal{C}$,
${{\boldsymbol{J}}_{ti}} = {}_s^c{\boldsymbol{R}}\left( {\begin{array}{*{20}{l}}
  {{\mathbb{I}_{3 \times 3}}}&-{{{\left[ {_e^s{\boldsymbol{R}}\Delta {{\boldsymbol{\rho }}_i}} \right]}_ \times }}
\end{array}} \right)$, and ${{\boldsymbol{J}}_e}$ is the jacobian matrix of the manipulator.
The selection matrix
${\boldsymbol{S}} = \left[ {\begin{array}{*{20}{c}}
{{{\boldsymbol{0}}_{6 \times 18}}}&{{{\mathbb{I}}_{6 \times 6}}}
\end{array}} \right]$
selects manipulator's joint to activate.
Therefore, the kinematics of
${{{\boldsymbol{s'}}}_{ti}}$  can be rewritten as
\begin{equation}
\label{eq:si_kinamtics2}
{{{\boldsymbol{\dot s'}}}_{ti}} = \frac{{{{\boldsymbol{\pi }}_{{{{{s'}}}_{ti}}}}}}{{r\left( {{{{\boldsymbol{Q'}}}_{ti}}} \right)}}{}_c^{c'}{{\boldsymbol{R}}^ \top }{{\boldsymbol{J}}_{ti}}{{\boldsymbol{J}}_e}{\boldsymbol{S}\boldsymbol{v}} - {\left[ {{\boldsymbol{\Omega }}_c^c} \right]_ \times }{{{\boldsymbol{s'}}}_{ti}}.
\end{equation}

Similar to the ${{\boldsymbol{h}}_o}$, using the centroid technology for $m$ redefined virtual points, we have
\begin{subequations}
\label{eq:ht_ht_dot}
\begin{align}
\label{eq:ht}
&{{\boldsymbol{h}}_t} = \frac{1}{m}\sum\limits_{i = 1}^m {{{{\boldsymbol{s'}}}_{ti}}},\, {\boldsymbol{L}_t} = \frac{1}{m}\sum\limits_{i = 1}^m \frac{ {\boldsymbol{\pi }}_{{s^\prime }_{ti}}}{| {{{{\boldsymbol{Q'}}}_{ti}}} |} \\
\label{eq:ht_dot}
&{{{\boldsymbol{\dot h}}}_t} =  - {\left[ {{\boldsymbol{\Omega }}_c^c} \right]_ \times }{{\boldsymbol{h}}_t} + {\boldsymbol{L}_t}{}_c^{c'}{\boldsymbol{R}^ \top }{{\boldsymbol{J}}_t}{{\boldsymbol{J}}_e}{\boldsymbol{Sv}}
\end{align}
\end{subequations}
%
where ${\boldsymbol{L}_t} \in {\mathbb{R}^{3 \times 3}}$ and ${{\boldsymbol{J}}_t} = \frac{1}{m}\sum\limits_{i = 1}^m {{{\boldsymbol{J}}_{ti}}}$. Notes that $| {{{{\boldsymbol{Q'}}}_{ti}}} |$  can
be calculated using the kinematics of the manipulator, and thus $\boldsymbol{L}_t$  is a known parameter.

With the target image and virtual point centroid position information, the visual error can be built as ${\boldsymbol{e}} = {{\boldsymbol{h}}_o} - {{\boldsymbol{h}}_t}$.
Differentiating the error $\boldsymbol{e}$  w.r.t time and combining  \eqref{eq:ho_dot} and \eqref{eq:ht_dot}, we have
\begin{equation}
\label{eq:error_dot}
{\boldsymbol{\dot e}} =  - {\left[ {{\boldsymbol{\Omega }}_c^c} \right]_ \times }{\boldsymbol{e}} - {\boldsymbol{L}_o}({\boldsymbol{v}}_c^c - {\boldsymbol{v}}_T^c) - {\boldsymbol{L}_t}{}_c^{c'}{\boldsymbol{R}^ \top }{{\boldsymbol{J}}_t}{{\boldsymbol{J}}_e}{\boldsymbol{Sv}}.
\end{equation}
The equation satisfies the passive-like property \cite{Zhong2020}, which separates linear velocity from angular velocity, and can be used to avoid controlling the attitude of the quadruped platform.
If ${\boldsymbol{L}_o}$ can be replaced by ${\boldsymbol{L}_t}$ and ${\boldsymbol{v}}_T^c$ can be estimated by a robust observer,
then according to \eqref{eq:error_dot}, the control input $\boldsymbol{v}$ that is independent of the target depth can be designed.

\subsection{Multivariable Super-Twisting Velocity Observer}
\label{sec:vel_observer}
To track the dynamic target, the linear velocity of the target needs to be observed.
However, since the target depth is unknown, ${{\boldsymbol{L}}_o}$ is an unknown parameter,
which brings difficulties to the observation.
As mentioned above, when the manipulator successfully grabs the target,  ${{\boldsymbol{L}}_o}$ will converge to ${{\boldsymbol{L}}_t}$. Meanwhile, ${{\boldsymbol{L}}_t}$ can be obtained from the kinematics of the manipulator and is a known parameter. Therefore, Let us assume that the gain ${\boldsymbol{0}} < {{\boldsymbol{L}}_o} < {{\boldsymbol{L}}_t}$ is always satisfied with some known constant ${{\boldsymbol{L}}_t}$. To exactly estimate the unknown velocity ${\boldsymbol{v}}_T^c$, let us design a robust observer based on the robust super-twisting observer (STO)\cite{Yang2022}:
\begin{subequations}
\label{eq:STO}
\begin{align}
\label{eq:STA1}
&\dot {\hat{\boldsymbol{h}}}_{o} =  - {\left[ {{\boldsymbol{\Omega }}_c^c} \right]_ \times }{{\boldsymbol{h}}_o} - {{\boldsymbol{L}}_t}{\boldsymbol{v}}_c^c + {k_1}{\varphi _1}({{\boldsymbol{e}}_o}){{\boldsymbol{e}}_o} + {{\boldsymbol{L}}_t}{\boldsymbol{y}} \\
&{\boldsymbol{\dot y}} = {k_2}{\varphi _2}({\boldsymbol{e}_o}){\boldsymbol{e}_o},{\ }{\varphi _1}({\boldsymbol{e}_o}) = {k_3}{\left| {\boldsymbol{e}_o} \right|^{ - p}} + {k_4}\\
&{\varphi _2}({\boldsymbol{e}_o}) = \left( {{k_3}(1 - p){{\left| {\boldsymbol{e}_o} \right|}^{ - p}} + {k_4}} \right){\varphi _1}({\boldsymbol{e}_o})
\end{align}
\end{subequations}
where ${\boldsymbol{e}_o} = {{\boldsymbol{h}}_o} - {\hat{\boldsymbol{h}}}_{o}$ is the estimate error, $0 < p \le 0.5$, and
${k_i} > 0,i = 1,2,3,4$ are the gains. With proper gain $k_i$ and the assumption ${\boldsymbol{0}} < {{\boldsymbol{L}}_o} < {{\boldsymbol{L}}_t}$, the variable ${\boldsymbol{y}}$ is an exact estimation of ${\boldsymbol{v}}_T^c$ within finite time.

\subsection{Model Predictive Control} \label{sec:mpc}
Based on the error dynamic \eqref{eq:error_dot}, the reference trajectory that guarantees error convergence can be generated as
\begin{subequations}
\label{eq:desired_traj}
\begin{align}
&{{\boldsymbol{v}}_B^d} = {}^I_c{{\boldsymbol{R}}}{({{\boldsymbol{K}}_b}{\boldsymbol{e}} + {\boldsymbol{v}}_T^c)} + {}^I_B{{\boldsymbol{R}}}{\left[ {{}^B_c{{\boldsymbol{t}}}} \right]_ \times }\boldsymbol{{\omega}_B} \\
&{\boldsymbol{\dot q}}_{arm}^d = {{\boldsymbol{K}}_a}{({{\boldsymbol{L}}_t}_c^{c'}{{\boldsymbol{R}}^ \top }{{\boldsymbol{J}}_t}{{\boldsymbol{J}}_e})^{\dagger}}{\boldsymbol{e}}
\end{align}
\end{subequations}
where ${\boldsymbol{e}}$ is the visual error represented in \eqref{eq:error_dot}, ${{\boldsymbol{K}}_b}$,  ${{\boldsymbol{K}}_a}$ are diagonal positive gain matrix, ${{}^B_c{{\boldsymbol{t}}}}$ is the offset vector between the camera frame $\cal{C}$ and the body frame $\cal{B}$,
$\boldsymbol{\dot q}_{arm}$ is the block vector of $\boldsymbol{v}$ that represents the joint velocity of the manipulator,
and ${\left(  \cdot  \right)^\dag }$ is the pseudo-inverse operator.
Because ${\boldsymbol{L}_o}$ and ${\boldsymbol{L}_t}$ are positive definite matrices, \eqref{eq:desired_traj} can make the visual error converge.
To simplify the controller, the desired angular velocity of the quadruped
platform keeps as zero.
For the attitude control, we can design an independent control law by constructing a visual feature only related to the attitude.
For example, a visual feature based on image moment is constructed in \cite{Zheng2017}.
With the desired linear and angular velocity, the desired position and Euler angles can be obtained by integrating.
Therefore, the reference trajectory of the quadruped platform is ${{\boldsymbol{x}}^{ref}} = {\left[ {\begin{array}{*{20}{l}}
  {{{\boldsymbol{\Phi }}^d_B}^ \top }&{{{\boldsymbol{p}}^d_B}^ \top }&{{{\boldsymbol{\omega }}^d_B}^ \top }&{{{\boldsymbol{v}}^d_B}^ \top }
\end{array}} \right]^ \top }$.

Following the work in \cite{DiCarlo2018}, here, a dynamics model of single rigid body is used to generate ground reaction forces (GRFs) of the quadruped manipulator.
Since the mass of the legs and manipulator only accounts for $20\%$ of the total mass in our robot, we ignore the motion of the legs and manipulator when calculating GRFs.
The difference from \cite{DiCarlo2018} is that the mass of the system is the sum of the masses of  the quadruped platform and manipulator, and the inertia of the system is the quadruped base's inertia adding the manipulator nominal inertia.
Therefore, GRFs can be calculated from MPC. The detailed construction and solution of MPC can refer to \cite{DiCarlo2018}.

The controller of legs uses a feedback law of PD control compensated with one feedforward term to compute joint torques:
\begin{flalign}\label{eq:leg_control}
 & {\boldsymbol{\tau }}_{leg,i} = {\boldsymbol{J}}_i^ \top [ {\boldsymbol{K}}_p\left( {{}_{\cal B}{{\boldsymbol{p}}_{i,{\rm{ref}}}} - {}_{\cal B}{{\boldsymbol{p}}_i}} \right) +\nonumber\\
 &\qquad  {\boldsymbol{K}}_d({{}_{\cal B}{{\boldsymbol{v}}_{i,{\rm{ref}}}} - {}_{\cal B}{{\boldsymbol{v}}_i}})  ]+ {\boldsymbol{\tau }}_{i,{\rm{ff}}}
\end{flalign}
where, ${}_{\cal B}{{\boldsymbol{p}}_i},{}_{\cal B}{{\boldsymbol{v}}_i} \in {\mathbb{R}^{{3}}}$ are the position and velocity of the $i$-$th$ foot, generated by Raibert heuristic\cite{Raibert1986},
${\boldsymbol{J}}_i$ is the foot Jacobian, $\boldsymbol{K}_p, \boldsymbol{K}_d$ are
diagonal positive matrix, and ${{\boldsymbol{\tau }}_{i,{\rm{ff}}}}$ is the feedforward torque.
For stance legs,  the feedforward torque can be calculated as
\begin{equation}\label{eq:stance_leg_control}
{{\boldsymbol{\tau }}_{i,\rm{ff}}} = {\boldsymbol{J}}_i^ \top {}_B^I{\boldsymbol{R}}^ \top {{\boldsymbol{f}}_i}
\end{equation}
where ${}_B^I{\boldsymbol{R}}$ represents the rotation matrix of the body frame $\cal B$ w.r.t the inertial coordinate frame $\cal I$, and ${{\boldsymbol{f}}_i}$ is the GRFs calculated by MPC.
For swing legs, the feedforward torque can be calculated as
\begin{equation}\label{eq:swing_leg_control}
{{\boldsymbol{\tau }}_{i,{\rm{ff}}}} = {{\boldsymbol{J}_{i}}^ \top }{{\boldsymbol{\Lambda }}_i}\left( {_{\cal B}{{\boldsymbol{a}}_{i,{\rm{ref}}}} - {{{\boldsymbol{\dot J}}}_i}{{{\boldsymbol{\dot q}}}_{leg,i}}} \right) + {{\boldsymbol{n}}_{leg,i}}
\end{equation}
where $\mathbf{\Lambda}_i \in \mathbb{R}^{3 \times 3}$  is the operational space inertia matrix, $_{\cal{B}}{\boldsymbol{a}_{i,\rm{ref}}} \in \mathbb{R}^3$ is the reference acceleration in the body frame, ${{\boldsymbol{\dot q}}_{leg,i}} \in {\mathbb{R}^3}$ is the vector of leg joint velocities, ${{\boldsymbol{n}}_{leg,i}}$ is the nonlinear effects (e.g., Coriolis, centrifugal and gravitational terms).

In manipulator control, we follow the work \cite{Rehman2016a} to consider the coupling of the manipulator to the quadruped platform.
The difference from \cite{Rehman2016a} is that we ignore the motion of the legs, and only consider the effect of quadruped base motion on the manipulator.
From \cite{Rehman2016a}, we can get
\begin{equation}\label{eq:arm_control_3}
{{\boldsymbol{M}}^{fl}}{{{\boldsymbol{\ddot q}}}_{arm}} + {{\boldsymbol{n}}^{fl}} = {{\boldsymbol{\tau }}_{arm}}
\end{equation}
where ${{\boldsymbol{M}}^{fl}} = {{\boldsymbol{M}}_{arm}} - {{\boldsymbol{F}}^T}{\left( {{{\boldsymbol{M}}_B}} \right)^{ - 1}}{\boldsymbol{F}}$,
${{\mathbf{n}}^{fl}} = {{\boldsymbol{n}}_{arm}} - {{\boldsymbol{F}}^T}{\left( {{{\boldsymbol{M}}_B}} \right)^{ - 1}}{{\boldsymbol{n}}_B}$,
${{{\boldsymbol{M}}_B}}$ is the rigid body inertia of the quadruped's base,
${{{\boldsymbol{F}}}}$ is the block matrix that encodes inertial coupling between the manipulator and the base,
${{{\boldsymbol{M}}_{arm}}}$ is the inertia of the manipulator, and
${{{\boldsymbol{n}}}}$ is the nonlinear effects (e.g., Coriolis, centrifugal and gravitational terms).
Therefore, the control law of manipulator can be designed as:
\begin{equation}\label{eq:arm_control_3}
{{{\boldsymbol{\ddot q}}}_{arm}} = {\boldsymbol{\ddot q}}_{arm}^d + {{\boldsymbol{K}}_{{d_{arm}}}}{{{\boldsymbol{\dot e}}}_{arm}} + {{\boldsymbol{K}}_{p_{arm}}}{{\boldsymbol{e}}_{arm}}
\end{equation}
where ${{{\boldsymbol{\ddot q}}}^d_{arm}}$ is obtained by differentiating ${\boldsymbol{\dot q}}_{arm}^d$, $\boldsymbol{K}_{d_{arm}}$, $\boldsymbol{K}_{p_{arm}}$ are diagonal positive matrix, and ${{\boldsymbol{e}}_{arm}}$, ${{{\boldsymbol{\dot e}}}_{arm}}$ are the tracking error between the actual and the desired joint positions/velocites, respectively.

\begin{figure*}[tp]
	\centering
	\subfigure[]{
        \label{fig:sub_observer_a}
		\includegraphics[width=0.305\linewidth]{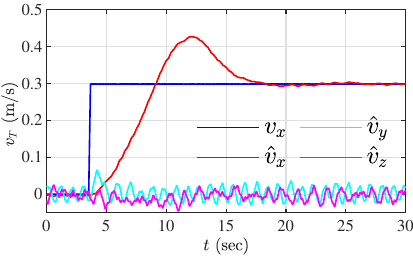}}
	\subfigure[]{
        \label{fig:sub_observer_b}
		\includegraphics[width=0.305\linewidth]{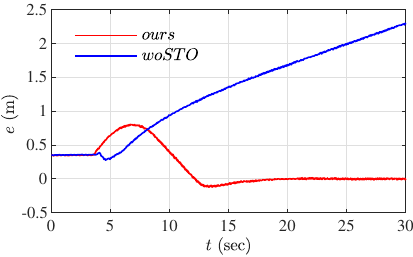}}
	\subfigure[]{
        \label{fig:sub_observer_c}
		\includegraphics[width=0.295\linewidth]{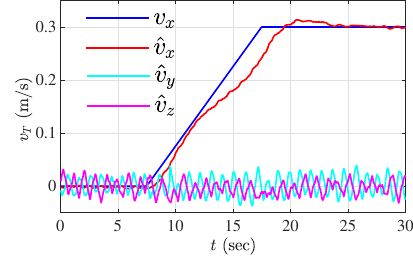}}
	\caption{Simulation results of tracking a dynamic target moving along with a straight line trajectory.
        (a) Evolution of the observed velocity with unknown $0.3$m/s, where
        $v_x$ is the $x$-component linear velocity of the target.
        $({\hat{v}_x}, {\hat{v}_y}, {\hat{v}_z})$ is the result of the velocity observer.
        (b) Time history of the tracking error between the manipulator's end-effector and the target,
        where the red line is the tracking error caused by our approach, and the blue line is the tracking error caused by the ``woSTO", denoting the approach \cite{Zhong2020} without using the STO \eqref{eq:STO}.
        (c) Evolution of the observing velocity with $0.03$m/s$^2$ acceleration.
        }
        \label{fig:ex_observer}
\end{figure*}

\begin{figure*}[tp]
	\centering
	\subfigure[]{
        \label{fig:sub_s_a}
		\includegraphics[width=0.305\linewidth]{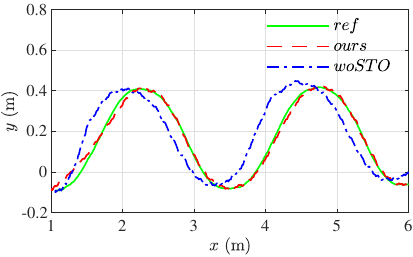}}
	\subfigure[]{
        \label{fig:sub_s_b}
		\includegraphics[width=0.305\linewidth]{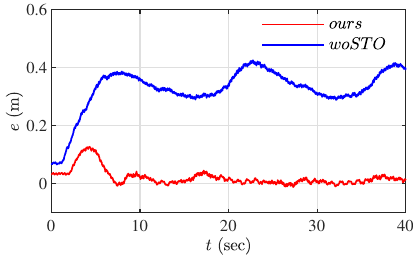}} \
	\subfigure[]{
        \label{fig:sub_s_c}
		\includegraphics[width=0.305\linewidth]{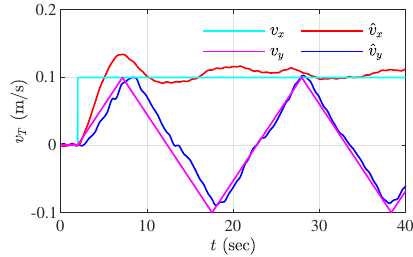}}
	\caption{Simulation results of tracking a dynamic target moving along with an unknown $S$-shaped trajectory.
        (a) Evolution of the 2D tracking trajectories, where the green solid line is the reference trajectory of the target,
        the red dotted line is the tracking trajectory caused by our approach, and the blue dot-and-dash line is the tracking trajectory produced by the method ``woSTO", denoting the approach
\cite{Zhong2020} without using the STO \eqref{eq:STO}.
        (b) Time history of the tracking error between the manipulator's end-effector and the dynamic target.
        (c) Evolution of the observed velocity with a $S$-shaped trajectory.
        }
        \label{fig:ex_s}
\end{figure*}
\section{EXPERIMENTS}
Various validations have been conducted both in simulation and experiments on the quadruped manipulator, as shown in Fig.~\ref{fig:qm_grasp}.
The quadruped manipulator consists of a torque-controllable quadrupedal robot, Aliengo\cite{wang2020unitree},  a 6-DOF manipulator, Kinova gen2\cite{Campeau-Lecours2018}, and the realsense D435i\cite{Grunnet-Jepsen2021} RGB-D camera from Intel.
The manipulator is lightweight ($4.4 $ kg), and allows for torque control of all six actuators.
It should be noted that only the RGB images from the RGB-D camera have been used in both simulations and experiments.
The MPC controller (Sec. \ref{sec:mpc}) runs on the user's computer (Intel Core i7-11700F@ 2.50GHz), and relies on a state estimator running at $400$Hz,
as shown in Fig. \ref{fig:qm_control}.
The state estimator used in the controller is based on the work in \cite{Bloesch2013}
to get an estimate of the base pose, linear and angular velocity by fusing  measurements from
the motors' encodes and IMU.
We use the open-source Pinocchio \cite{Carpentier2019} to generate the model of the kinematics and dynamics of the robot.
The MPC (Sec. \ref{sec:mpc}) is solved by QPOASES \cite{Ferreau2014}, and runs at a frequency of approximately $100$Hz.
All of the examples discussed in this paper are supported by the video submission\footnote{Available at \url{https://youtu.be/Tep_d-BOPwo}}.

\subsection{Simulation Results for Dynamic Target Tracking}
The primary purpose of this experiment is to demonstrate the effectiveness of the proposed approach.
The proposed approach was validated in a simulator developed in the Gazebo environment \cite{Koenig2004}.
The quadruped was under the trot gait.
Feature points ware provided by a visual markers' plane, and the target was located $0.15$m above the visual markers' plane (see the attached video).
Opencv\cite{Bradski2000} was used to detect target feature points $\boldsymbol{q}_{o i} \eqref{eq:SphericalProjection}$.
The target was set to unknown constant and changing velocity, respectively.
The trajectory of the target is set as a straight line or an $S$-shaped curve
to verify the effectiveness in different directions.
The robot's task goal was to track the dynamic tracking point accurately.
The parameters of STO \eqref{eq:STO} are $p=0.4$, $k_1=10$, $k_2=100$, $k_3=0.05$, $k_4=0.05$.
At the same time, an IBVS based on spherical projection (without STO)\cite{Zhong2020} was used as the comparison.

\subsubsection{Straight line with unknown velocity}
Firstly, the robot tracked the target with a straight line motion.
The target's velocity was set to $0.3$m/s and $0.5$m/s for uniform motion, respectively.
The tracking results are shown in Fig. \ref{fig:sub_observer_a} and Fig. \ref{fig:sub_observer_b}.
In Fig. \ref{fig:sub_observer_a}, $v_x$ is the target's velocity expressed in the inertial coordinate frame $\cal I$, and $({\hat{v}_x}, {\hat{v}_y}, {\hat{v}_z})$ is the result of the STO \eqref{eq:STO}.
It can be observed from Fig. \ref{fig:sub_observer_a} that our approach can accurately estimate the velocity of the target.
Fig. \ref{fig:sub_observer_b} shows the tracking error between the manipulator's end-effector and the target.
It shows that the robot can accurately track the target with the unknown velocity with our approach, while the method in \cite{Zhong2020} failed to work.
The attached video shows the same result when the target velocity was set to $0.5$m/s.
The  method in \cite{Zhong2020} failed to track because the unknown target velocity made the visual error diverge and eventually, causing the target to disappear from the camera's field of view.
These results illustrate the efficiency of the proposed approach for tracking the dynamic object.

\subsubsection{Straight line with changing velocity}
Then, the target was set to move along with a straight line with changed velocities.
The acceleration was set to $0.015$m/s$^2$ and $0.03$m/s$^2$, respectively, and the maximum velocity was set to $0.3$m/s.
The results of tracking the target moving with $0.03$m/s$^2$ are shown in Fig. \ref{fig:sub_observer_c}.
From Fig. \ref{fig:sub_observer_c}, we can see that the robot can accurately estimate the changing velocity of the target.

\begin{figure*}[htp]
\centering
\includegraphics[scale=0.98]{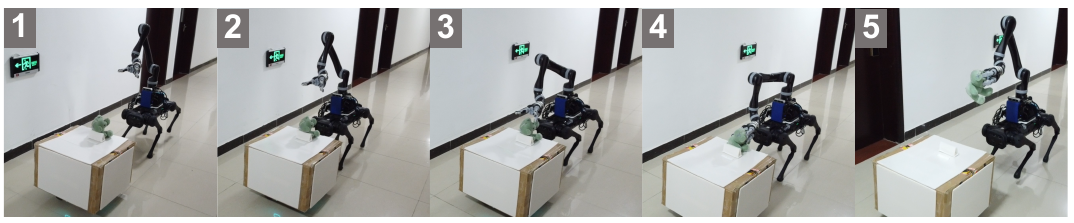}
\caption{Some snapshots of the experiments of grasping a dynamic target. The real robot grasped a dynamic object moving with an unknown velocity of $0.1$m/s. Feature points are obtained by the 2D detection network (YOLO). }
\label{fig:ex_snap}
\end{figure*}

\subsubsection{$S$-shaped trajectory}
Based on the straight-line motions, the $S$-shaped motion was added to further demonstrate the effectiveness
of our approach in different directions.
The target's velocity in the $x$ direction was set to $0.1$m/s, the acceleration in the $y$ direction was set to $0.02$ m/s$^2$, and the maximum velocity in the $y$ direction was set to $0.1$m/s. The results are shown in Fig. \ref{fig:ex_s}, where Fig. \ref{fig:sub_s_a} shows the 2-D trajectories of the target and the manipulator's end-effector with different tracking approaches.
Fig. \ref{fig:sub_s_b} shows the tracking error between the manipulator's end-effector and the target.
Fig. \ref{fig:sub_s_c} shows results of estimated velocities.
These results show that by effectively estimating the unknown velocities of the dynamic target through the proposed velocity observer, the robot can track the $S$-shaped trajectory of the target with the smaller error.

\begin{figure}[htp]
	\centering
	\subfigure[]{
    \label{fig:sub_grasping_a}
		\includegraphics[width=0.47\linewidth]{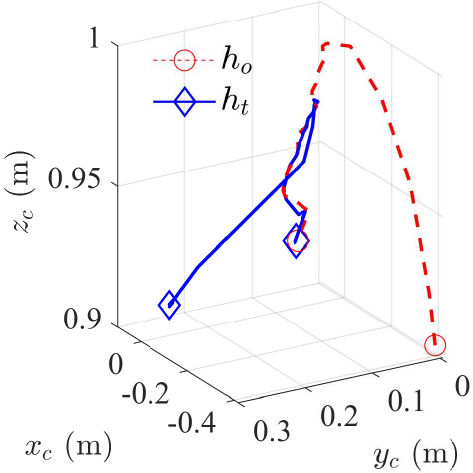}}
	\subfigure[]{
    \label{fig:sub_grasping_b}
		\includegraphics[width=0.47\linewidth]{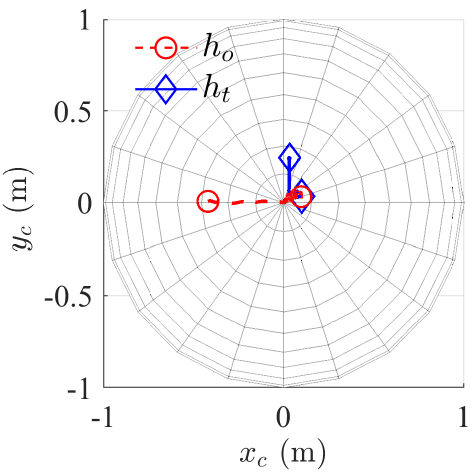}} \\

    \subfigure[]{
        \label{fig:sub_grasping_c}
		\includegraphics[width=0.47\linewidth]{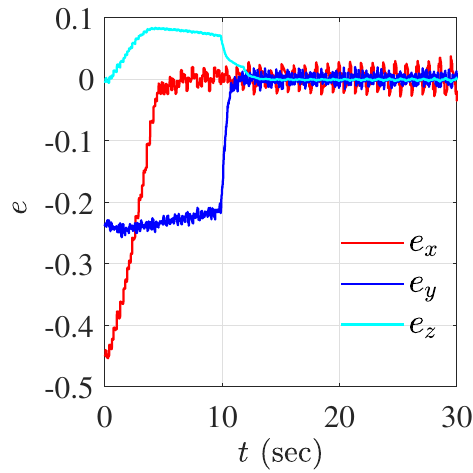}}
	\subfigure[]{
        \label{fig:sub_grasping_d}
		\includegraphics[width=0.47\linewidth]{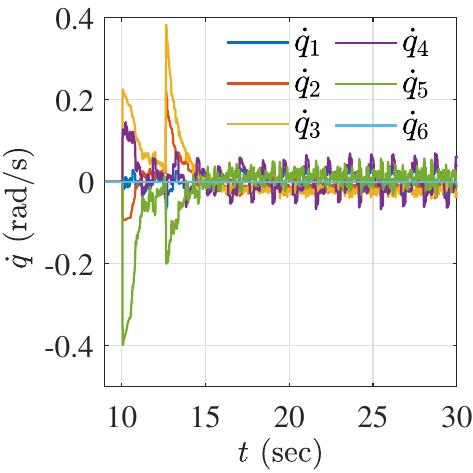}} \\

    \subfigure[]{
        \label{fig:sub_grasping_e}
		\includegraphics[width=0.47\linewidth]{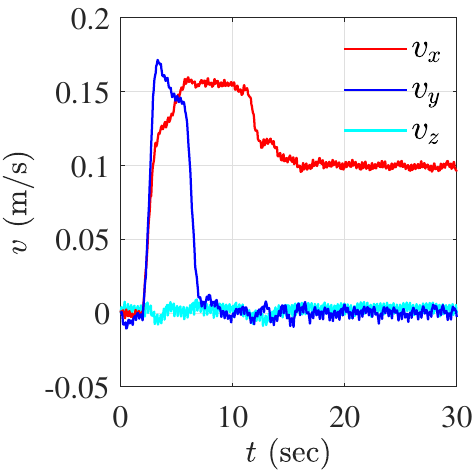}}
    \subfigure[]{
        \label{fig:sub_grasping_f}
		\includegraphics[width=0.47\linewidth]{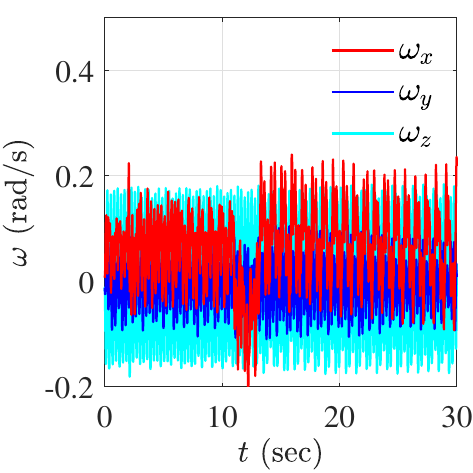}}
	\caption{The experiments of grasping the dynamic target moving with unknown velocities.
    (a) and (b) Evolution of the target and virtual centroid point $({{\boldsymbol{h}}_o}, {{\boldsymbol{h}}_t})$ in the spherical image space, where the red dotted line is the target centroid point $\boldsymbol{h}_o$, and the blue solid line is the virtual centroid point $\boldsymbol{h}_t$.
    (c) Evolution of the visual error $\boldsymbol{e}$.
    (d) Time history of the manipulator joint velocity.
    (e) Time history of the quadruped linear velocity in the inertial frame.
    (f) Time history of the quadruped angular velocity. }
    \label{fig:ex_grasping}
\end{figure}

\subsection{Real World Results for Grasping Dynamic Target}
To illustrate the efficiency of our approach of tracking various targets for the quadruped manipulator in real world, we combined a 2D detection network (YOLO\cite{Redmon2016}) to grab a household object (teddy bear).
The target's velocity was set to $0$m/s and $0.1$m/s, respectively (see the attached video).
The robot's task is to grasp the dynamic target moving with these unknown velocities.
In the task, feature points $\boldsymbol{q}_{o i}$ \eqref{eq:SphericalProjection} were built with the corner points of the target bounding box obtained by YOLO. Then, with the formula \eqref{eq:SphericalProjection}, \eqref{eq:ho_ho_dot}, \eqref{eq:ht_ht_dot} and the STO \eqref{eq:STO}, the reference trajectory can be generate by \eqref{eq:desired_traj}. Finally the robot is controlled by MPC (Sec. \ref{sec:mpc}).

When the target is static, the average of grasping error is about $0.03$m and the success rate grasps is about $96\%$, while when the target is moving with $0.1$m/s, the average of grasping error is about $0.06$m and the success rate grasps is $82\%$.
The quantitative measurements were based on $50$ experiments, where the grasping error is the average of Euclidean distances between the end-effector and the center of the target when the quadruped manipulator successfully grasps.
The results show that our approach has a high grasping success rate and low grasping error under different velocities of dynamic target.
The attached video shows that our approach can make the robot to produce coordinated motion, where
the quadruped is constantly adjusting its position to help the manipulator successfully grab the target.

Some snapshots in Fig. \ref{fig:ex_snap} show the results at various moments when a dynamic object moving with $0.1$m/s, and the detailed results are shown in Fig. \ref{fig:ex_grasping}.
To avoid the manipulator getting into a singularity when the target is too far away, the manipulator is only controlled when the target is in the workspace of the manipulator.
Thus, the manipulator is controlled after $10s$, as shown in Fig.\ref{fig:sub_grasping_d}. $\boldsymbol{\dot{q}}$ is the joint velocity of the manipulator.
Fig.\ref{fig:sub_grasping_f} and Fig.\ref{fig:sub_grasping_c} show the angular velocity of the quadruped platform and the visual error $\boldsymbol{e}$, respectively.
We can see that although the angular velocity of the quadruped platform has been changing, it has not affected the convergence of the visual error.
The results illustrate that the effectiveness of the passive-like visual equation \eqref{eq:error_dot} of our approach.
Fig.\ref{fig:sub_grasping_e} shows the quadruped platform's  linear velocity in the inertial coordinate frame $\cal{I}$.
The results show that the velocity eventually converges to the  approximate target's velocity $(0.1, 0.0, 0.0) $m/s, which is estimated by STO \eqref{eq:STO}.
Fig. \ref{fig:sub_grasping_a} and Fig. \ref{fig:sub_grasping_b} show  the evolution of the target and virtual centroid point $({{\boldsymbol{h}}_o}, {{\boldsymbol{h}}_t})$ in the spherical image space.
The evolution process of the target centroid ${{\boldsymbol{h}}_o}$ is represented by a red dotted line, and the virtual centroid ${{\boldsymbol{h}}_t}$ is represented by a blue solid line.
From the figure, the target centroid ${{\boldsymbol{h}}_o}$ and the virtual centroid ${{\boldsymbol{h}}_t}$ eventually tend to coincide.
Since ${{\boldsymbol{h}}_o}$ and ${{\boldsymbol{h}}_t}$ can represent the center of the target and the manipulator's end-effector respectively, when ${{\boldsymbol{h}}_o}$  coincides with ${{\boldsymbol{h}}_t}$, the origin of the manipulator's end-effector $O_e$ coincides with the target origin $O_T$ in cartesian space.
The results demonstrate the effectiveness and robustness of our approach to dynamic target grasping.
The above results show the approach's effectiveness in improving the autonomy of the quadruped manipulator.


\section{CONCLUSIONS}
This paper proposed a novel spherical image-based approach that enable the quadruped manipulator to track a dynamic object of moving with an unknown velocity. In contrast to the conventional methods, the new approach can exactly estimate the unknown velocity with only an onboard RGB camera without requiring any marker board or depth estimation.
Moreover, the passive-like property of the visual error is exploited to eliminate the influence of the under-actuated angular velocity of the quadruped platform, which enables the manipulator to track the dynamic object with simultaneous locomotion.
The experiments demonstrate that our approach can robustly estimate the target's unknown constant or changing velocity and track dynamic targets.
Combined with 2D detection methods, our method can be used in various application scenarios, which greatly enhances the autonomy of the quadruped manipulator.
Future work may extend the application of the proposed IBVS to tracking targets with angular velocity.

\addtolength{\textheight}{-12cm}   



%

%


%
%
%
\bibliographystyle{IEEEtran}
\bibliography{ref}

\begin{thebibliography}{10}
\providecommand{\url}[1]{#1}
\csname url@samestyle\endcsname
\providecommand{\newblock}{\relax}
\providecommand{\bibinfo}[2]{#2}
\providecommand{\BIBentrySTDinterwordspacing}{\spaceskip=0pt\relax}
\providecommand{\BIBentryALTinterwordstretchfactor}{4}
\providecommand{\BIBentryALTinterwordspacing}{\spaceskip=\fontdimen2\font plus
\BIBentryALTinterwordstretchfactor\fontdimen3\font minus
  \fontdimen4\font\relax}
\providecommand{\BIBforeignlanguage}[2]{{%
\expandafter\ifx\csname l@#1\endcsname\relax
\typeout{** WARNING: IEEEtran.bst: No hyphenation pattern has been}%
\typeout{** loaded for the language `#1'. Using the pattern for}%
\typeout{** the default language instead.}%
\else
\language=\csname l@#1\endcsname
\fi
#2}}
\providecommand{\BIBdecl}{\relax}
\BIBdecl

\bibitem{biswal2021development}
P.~Biswal and P.~K. Mohanty, ``Development of quadruped walking robots: A
  review,'' \emph{Ain Shams Engineering Journal}, vol.~12, no.~2, pp.
  2017--2031, 2021.

\bibitem{Li2011}
Y.~Li, B.~Li, J.~Ruan, and X.~Rong, ``Research of mammal bionic quadruped
  robots: A review,'' in \emph{2011 IEEE 5th International Conference on
  Robotics, Automation and Mechatronics (RAM)}.\hskip 1em plus 0.5em minus
  0.4em\relax IEEE, 2011, pp. 166--171.

\bibitem{Rehman2016a}
B.~U. Rehman, M.~Focchi, J.~Lee, H.~Dallali, D.~G. Caldwell, and C.~Semini,
  ``Towards a multi-legged mobile manipulator,'' in \emph{2016 IEEE
  International Conference on Robotics and Automation (ICRA)}.\hskip 1em plus
  0.5em minus 0.4em\relax IEEE, 2016, pp. 3618--3624.

\bibitem{Bellicoso2019a}
C. D. Bellicoso, K. Kr{\"a}mer, M. St{\"a}uble, D. Sako, F. Jenelten, M. Bjelonic, and M. Hutter,
``Alma-articulated locomotion and manipulation for a torque-controllable robot,'' in \emph{2019 International conference on
robotics and automation (ICRA)}.\hskip 1em plus 0.5em minus 0.4em\relax IEEE,
2019, pp. 8477--8483.

\bibitem{Ferrolho2020}
H.~Ferrolho, W.~Merkt, V.~Ivan, W.~Wolfslag, and S.~Vijayakumar, ``Optimizing
  dynamic trajectories for robustness to disturbances using polytopic
  projections,'' in \emph{2020 IEEE/RSJ International Conference on Intelligent
  Robots and Systems (IROS)}.\hskip 1em plus 0.5em minus 0.4em\relax IEEE,
  2020, pp. 7477--7484.

\bibitem{Sleiman2021}
J.-P. Sleiman, F.~Farshidian, M.~V. Minniti, and M.~Hutter, ``A unified mpc
  framework for whole-body dynamic locomotion and manipulation,'' \emph{IEEE
  Robotics and Automation Letters}, vol.~6, no.~3, pp. 4688--4695, 2021.

\bibitem{Chiu2022a}
J.-R. Chiu, J.-P. Sleiman, M.~Mittal, F.~Farshidian, and M.~Hutter, ``A
  collision-free mpc for whole-body dynamic locomotion and manipulation,'' in
  \emph{2022 International Conference on Robotics and Automation (ICRA)}.\hskip
  1em plus 0.5em minus 0.4em\relax IEEE, 2022, pp. 4686--4693.

\bibitem{zimmermann2021go}
S.~Zimmermann, R.~Poranne, and S.~Coros, ``Go fetch!-dynamic grasps using
  boston dynamics spot with external robotic arm,'' in \emph{2021 IEEE
  International Conference on Robotics and Automation (ICRA)}.\hskip 1em plus
  0.5em minus 0.4em\relax IEEE, 2021, pp. 4488--4494.

\bibitem{Mittal2021}
M.~Mittal, D.~Hoeller, F.~Farshidian, M.~Hutter, and A.~Garg, ``Articulated
  object interaction in unknown scenes with whole-body mobile manipulation,''
  in \emph{2022 IEEE/RSJ International Conference on Intelligent Robots and
  Systems (IROS)}.\hskip 1em plus 0.5em minus 0.4em\relax IEEE, 2022, pp.
  1647--1654.

\bibitem{Ming2021}
Y.~Ming, X.~Meng, C.~Fan, and H.~Yu, ``Deep learning for monocular depth
  estimation: A review,'' \emph{Neurocomputing}, vol. 438, pp. 14--33, 2021.

\bibitem{Rafique2020}
M.~A. Rafique and A.~F. Lynch, ``Output-feedback image-based visual servoing
  for multirotor unmanned aerial vehicle line following,'' \emph{IEEE
  Transactions on Aerospace and Electronic Systems}, vol.~56, no.~4, pp.
  3182--3196, 2020.

\bibitem{Ye2015}
W.~Ye, Z.~Li, C.~Yang, J.~Sun, C.-Y. Su, and R.~Lu, ``Vision-based human
  tracking control of a wheeled inverted pendulum robot,'' \emph{IEEE
  transactions on cybernetics}, vol.~46, no.~11, pp. 2423--2434, 2015.

\bibitem{kolter2009stereo}
J.~Z. Kolter, Y.~Kim, and A.~Y. Ng, ``Stereo vision and terrain modeling for
  quadruped robots,'' in \emph{2009 IEEE International Conference on Robotics
  and Automation}.\hskip 1em plus 0.5em minus 0.4em\relax IEEE, 2009, pp.
  1557--1564.

\bibitem{Chen2019}
F.~Chen, M.~Selvaggio, and D.~G. Caldwell, ``Dexterous grasping by
  manipulability selection for mobile manipulator with visual guidance,''
  \emph{IEEE Transactions on Industrial Informatics}, vol.~15, no.~2, pp.
  1202--1210, 2018.

\bibitem{Hamel2002}
T.~Hamel and R.~Mahony, ``Visual servoing of an under-actuated dynamic
  rigid-body system: an image-based approach,'' \emph{IEEE Transactions on
  Robotics and Automation}, vol.~18, no.~2, pp. 187--198, 2002.

\bibitem{Odile2007a}
O.~Bourquardez, R.~Mahony, N.~Guenard, F.~Chaumette, T.~Hamel, and L.~Eck,
  ``Image-based visual servo control of the translation kinematics of a
  quadrotor aerial vehicle,'' \emph{IEEE Transactions on Robotics}, vol.~25,
  no.~3, pp. 743--749, 2009.

\bibitem{Guenard2008}
N.~Guenard, T.~Hamel, and R.~Mahony, ``A practical visual servo control for an
  unmanned aerial vehicle,'' \emph{IEEE Transactions on Robotics}, vol.~24,
  no.~2, pp. 331--340, 2008.

\bibitem{Zhong2020}
H.~Zhong, Z.~Miao, Y.~Wang, J.~Mao, L.~Li, H.~Zhang, Y.~Chen, and R.~Fierro,
  ``A practical visual servo control for aerial manipulation using a spherical
  projection model,'' \emph{IEEE Transactions on Industrial Electronics},
  vol.~67, no.~12, pp. 10\,564--10\,574, 2019.

\bibitem{Mahony2008}
\BIBentryALTinterwordspacing
R.~Mahony, P.~Corke, and T.~Hamel, ``{Dynamic Image-Based Visual Servo Control
  Using Centroid and Optic Flow Features},'' \emph{Journal of Dynamic Systems,
  Measurement, and Control}, vol. 130, no.~1, 12 2007, 011005. [Online].
  Available: \url{https://doi.org/10.1115/1.2807085}
\BIBentrySTDinterwordspacing

\bibitem{Herisse2012}
B.~Heriss{\'e}, T.~Hamel, R.~Mahony, and F.-X. Russotto, ``Landing a vtol
  unmanned aerial vehicle on a moving platform using optical flow,'' \emph{IEEE
  Transactions on robotics}, vol.~28, no.~1, pp. 77--89, 2011.

\bibitem{Serra2016}
P.~Serra, R.~Cunha, T.~Hamel, D.~Cabecinhas, and C.~Silvestre, ``Landing of a
  quadrotor on a moving target using dynamic image-based visual servo
  control,'' \emph{IEEE Transactions on Robotics}, vol.~32, no.~6, pp.
  1524--1535, 2016.

\bibitem{Yang2022}
X.~Yang, X.~Xiong, Z.~Zou, Y.~Lou, S.~Kamal, and J.~Li, ``Discrete-time
  multivariable super-twisting algorithm with semi-implicit euler method,''
  \emph{IEEE Transactions on Circuits and Systems II: Express Briefs}, vol.~69,
  no.~11, pp. 4443--4447, 2022.

\bibitem{DiCarlo2018}
J.~Di~Carlo, P.~M. Wensing, B.~Katz, G.~Bledt, and S.~Kim, ``Dynamic locomotion
  in the mit cheetah 3 through convex model-predictive control,'' in \emph{2018
  IEEE/RSJ international conference on intelligent robots and systems
  (IROS)}.\hskip 1em plus 0.5em minus 0.4em\relax IEEE, 2018, pp. 1--9.

\bibitem{Zheng2017}
D.~Zheng, H.~Wang, J.~Wang, S.~Chen, W.~Chen, and X.~Liang, ``Image-based
  visual servoing of a quadrotor using virtual camera approach,''
  \emph{IEEE/ASME Transactions on Mechatronics}, vol.~22, no.~2, pp. 972--982,
  2016.

\bibitem{Raibert1986}
M.~H. Raibert, \emph{Legged robots that balance}.\hskip 1em plus 0.5em minus
  0.4em\relax MIT press, 1986.

\bibitem{wang2020unitree}
X.~Wang, ``Unitree robotics,'' \emph{URL https://www. unitree. com}, 2020.

\bibitem{Campeau-Lecours2018}
A.~Campeau-Lecours, H.~Lamontagne, S.~Latour, P.~Fauteux, V.~Maheu, F.~Boucher,
  C.~Deguire, and L.-J.~C. L'Ecuyer, ``Kinova modular robot arms for service
  robotics applications,'' in \emph{Rapid Automation: Concepts, Methodologies,
  Tools, and Applications}.\hskip 1em plus 0.5em minus 0.4em\relax IGI global,
  2019, pp. 693--719.

\bibitem{Grunnet-Jepsen2021}
A.~Grunnet-Jepsen, J.~N. Sweetser, P.~Winer, A.~Takagi, and J.~Woodfill,
  ``Projectors for intel{\textregistered} realsense™ depth cameras d4xx,''
  \emph{Intel Support, Interl Corporation: Santa Clara, CA, USA}, 2018.

\bibitem{Bloesch2013}
M.~Bloesch, M.~Hutter, M.~A. Hoepflinger, S.~Leutenegger, C.~Gehring, C.~D.
  Remy, and R.~Siegwart, ``State estimation for legged robots-consistent fusion
  of leg kinematics and imu,'' \emph{Robotics}, vol.~17, pp. 17--24, 2013.

\bibitem{Carpentier2019}
J.~Carpentier, G.~Saurel, G.~Buondonno, J.~Mirabel, F.~Lamiraux, O.~Stasse, and
  N.~Mansard, ``The pinocchio c++ library: A fast and flexible implementation
  of rigid body dynamics algorithms and their analytical derivatives,'' in
  \emph{2019 IEEE/SICE International Symposium on System Integration
  (SII)}.\hskip 1em plus 0.5em minus 0.4em\relax IEEE, 2019, pp. 614--619.

\bibitem{Ferreau2014}
H.~J. Ferreau, C.~Kirches, A.~Potschka, H.~G. Bock, and M.~Diehl, ``qpoases: A
  parametric active-set algorithm for quadratic programming,''
  \emph{Mathematical Programming Computation}, vol.~6, pp. 327--363, 2014.

\bibitem{Koenig2004}
N.~Koenig and A.~Howard, ``Design and use paradigms for gazebo, an open-source
  multi-robot simulator,'' in \emph{2004 IEEE/RSJ International Conference on
  Intelligent Robots and Systems (IROS)(IEEE Cat. No. 04CH37566)},
  vol.~3.\hskip 1em plus 0.5em minus 0.4em\relax IEEE, 2004, pp. 2149--2154.

\bibitem{Bradski2000}
G.~Bradski, ``The opencv library.'' \emph{Dr. Dobb's Journal: Software Tools
  for the Professional Programmer}, vol.~25, no.~11, pp. 120--123, 2000.

\bibitem{Redmon2016}
J.~Redmon, S.~Divvala, R.~Girshick, and A.~Farhadi, ``You only look once:
  Unified, real-time object detection,'' in \emph{Proceedings of the IEEE
  conference on computer vision and pattern recognition}, 2016, pp. 779--788.

\end{thebibliography}

\end{document}